\documentclass[review,3p,times,sort&compress]{elsarticle}
\usepackage{appendix}
\usepackage{graphicx}
\usepackage{longtable}
\usepackage{tabularx}
\usepackage{xcolor}
\usepackage{hyperref}
\usepackage{booktabs}
\usepackage{subcaption}
\usepackage{cancel}
\usepackage{soul}

\usepackage[utf8]{inputenc}

\usepackage{lscape}
\usepackage{pdflscape}

\newcolumntype{C}[1]{>{\centering\arraybackslash}p{#1}}
\newcolumntype{L}[1]{>{\raggedright\arraybackslash}p{#1}}
\graphicspath{ {images/} }

\usepackage{makecell}

\usepackage{multirow}
\usepackage{longtable} 

\journal{Computers in Biology and Medicine}

\begin{document}

\begin{frontmatter}
%\title{An approach to select the most appropriate machine learning method for cell morphology analysis: case study for image classification of Sickle-cell Disease}
\title{Sickle-cell disease diagnosis support selecting the most appropriate machine learning method: Towards a general and interpretable approach for cell morphology analysis from microscopy images.}
\author[3]{Nataša Petrović}\ead{npe785@uib.es}
\author[3]{Gabriel Moyà-Alcover}\ead{gabriel.moya@uib.es}
\author[3]{Antoni Jaume-i-Capó\corref{cor1}}\ead{antoni.jaume@uib.es}
\author[1,2]{Manuel González-Hidalgo}\ead{manuel.gonzalez@uib.es}

\address[1]{SCOPIA research group. University of the Balearic Islands. Dpt. of Mathematics and Computer Science. 
Crta. Valldemossa, km 7.5, E-07122 Palma, Spain
}
\address[2]{Health Research Institute of the Balearic Islands (IdISBa), E-07010 Palma, Spain
}

\address[3]{UGiVIA research group. University of the Balearic Islands. Dpt. of Mathematics and Computer Science. 
Crta. Valldemossa, km 7.5, E-07122 Palma, Spain 
}

\cortext[cor1]{Corresponding author}

\begin{abstract}

%\noindent{\sl Background and Objective}: 
In this work we propose an approach to select the  classification method and features, based on the state-of-the-art, with best performance for diagnostic support through peripheral blood smear images of red blood cells. In our case we used samples of patients with sickle-cell disease which can be generalized for other study cases. To trust the behavior of the proposed system, we also analyzed the interpretability.
%\noindent{\sl Methods}: 
We pre-processed and segmented microscopic images, to ensure high feature quality. We applied the methods used in the literature to extract the features from blood cells and the machine learning methods to classify their morphology. Next, we searched for their best parameters from the resulting data in the feature extraction phase. Then, we found the best parameters for every classifier using Randomized and Grid search. 
%\noindent{\sl Results}:
%\noindent{\sl Conclusions}: 
For the sake of scientific progress, we published parameters for each classifier, the implemented code library, the confusion matrices with the raw data, and we used the public erythrocytesIDB dataset for validation. We also defined how to select the most important features for classification to decrease the complexity and the training time, and for interpretability purpose in opaque models. Finally, comparing the best performing classification methods with the state-of-the-art, we obtained better results even with interpretable model classifiers.
\end{abstract}


\end{frontmatter}

\section{Introduction}

%\subsection{SCD}
In sickle-cell Disease (SCD), red blood cells (RBCs) have the shape of a sickle or half-moon instead of the smooth, circular shape as normal cells have. Because of the irregular shape, the cells can be stuck in the blood vessels, which causes blockages of the blood flow in the body. Lack of elasticity and their irregular shape lead to shortened life of sickle erythrocytes and subsequent anaemia, often called sickle-cell anaemia. The WHO document "Sickle-cell disease and other haemoglobin disorders"~\cite{WHO} indicates that in 2011 around a 5\% of the world's population carries trait genes for hemoglobin disorders, mainly sickle cell disease and thalassemia. SCD is caused by inherited mutated haemoglobin genes from both parents. The document also indicates that the percentage of people who carry these genes is as high as 25\% in some regions and over 300,000 babies with severe haemoglobin disorders are born each year. SCD is spread among people whose ancestors are from sub-Saharan Africa, India, Saudi Arabia and Mediterranean countries. Around the world~\cite{abubakar2015global,naghavi2017global}, SCD caused 112,900 deaths in 1990, 176,200 deaths in 2013 and  55,3000 deaths in 2016. 

%\subsection{Treatment}
SCD cannot be cured but can be treated to ease the pain and prevent problems caused by this disease. One of the methods for diagnosing SCD is by observation of a patient's peripheral blood samples under a microscope and counting sickle cells, which is a tedious and time-consuming task.  Morphological analysis of peripheral blood smears is a crucial diagnostic aid because offers important information for diagnosis~\cite{merino2018optimizing}. Therefore, monitoring of these patients is an urgent requirement.

%\subsection{ML for medical image analysis}
Machine learning (ML) algorithms have been used for solving different problems for medical image analysis~\cite{zhu2017retinal,shi2016stacked}. ML techniques are also used for analysis of microscopic blood smear images. These techniques make easier and cheaper solving tasks of diagnosing the different blood diseases such as malaria and different types of anemia and leukemia~\cite{das2013machine, maity2017ensemble}. Therefore, SCD is a great candidate for using ML algorithms as diagnosis support, so the medical experts could have more time to solve more complicated tasks. 

%\subsection{State-of-the-art: Machine learning in the morphological analysis of RBCs}
A review about blood cells classification, focused on malignant lymphoid cells and blast cells, was published in~\cite{rodellar2018}. There are three main steps in image processing systems that are able to recognize different classes of RBC in peripheral blood smear images: image segmentation and cell detection, feature extraction and/or selection, and finally classification. The two first steps will be addressed in subsequent sections, in next paragraphs we survey the most popular RBCs classification methods for images of peripheral blood smear samples.

Automatic classification is the process that tries to assign a set of features or descriptors to a specific class among a set of known classes defined as target. This is a well-established problem and dealt with machine learning techniques~\cite{friedman2001elements}. In RBCs classification the issue is assigning a given cell image its corresponding group. 

Since the early work of Bacus et al.~\cite{bacus1976image}, much efforts has been made to design  automatic classification methods of RBCs based on images.  Among the most used classifiers for RBCs classification from peripheral blood smear samples, we found artificial neural networks (ANN), multilayer perceptron  (MLP), Naïve Bayes (NB), support vector machines (SVM),  k-nearest neighbors algorithm (kNN), and decision trees (DT). The most popular are, by far, the SVM, kNN and ANN classifiers.

One of the first ML methods used for the purpose previously described, was the classification and regression tree (CART) in 1994. That year, Wheeless et al.~\cite{wheeless1994classification} used CART to develop a cell classifier in three target classes (normal, sickle and other abnormal cells). A total of 42 features were extracted, and circularity was selected as the sole feature needed for segregating cells into the target classes. After segmentation, cells touching the borders or cluster cells were rejected, and individual detected cells were classified. The agreement between the automated classification and the classification provided by an expert was: 89\% for normal disocytes, 73\% for abnormal cells and 92\% for sickle cells. The CART classifier would appear later in 2012 in the work of Das et al.~\cite{das2012quantitative}. 

Maity et al.~\cite{maity2017ensemble} presented an ensemble rule-based decision-making approach for morphological classification of RBC. The decision about the abnormality was taken using multiple rule based expert systems. The  rules were generated using a C4.5 DT algorithm. The proposed ensemble approach classified the RBCs in eight types of abnormal erythrocytes using a set of 41 shape features. Acharya et al.~\cite{acharya2017} developed a decision system  with a set of rules to identify eleven categories of RBCs based on 8 geometrical categories. The authors used the K-medoids algorithm to obtain white blood cells (WBC) and granulometry analysis to separate the RBCs from WRCs. Clusters were not considered. 

Akrimi et al.~\cite{akrimi2014classification} used SVM to classify RBCs as either normal or abnormal for anemia diagnosis. Authors used an automated feature selection algorithm to obtain a set of 271 features, but they do not publish the details. Moreover, the kernel used in the experiments is not clear and the work is impossible to reproduce. The SVM classifier was used also in~\cite{lotfi2015detection,rodrigues2016morphological}.

The classification of RBCs using ANN were addressed in~\cite{veluchamy2012feature,tomari2014computer, lotfi2015detection,elsalamony2016detection}. After segmentation, Veluchamy et al.~\cite{veluchamy2012feature} used 27 features with an ANN (4 geometrical, 16 statistical and 7 moment invariant) to classify RBCs as normal or abnormal. Tomari et al.~\cite{tomari2014computer} used geometrical properties of cell (such us perimeter, area and seven Hu moment features) in combination with an ANN to classify individuals RBCs as normal or abnormal. Full range images were considered but clusters were not analysed in these works. Elsalamony~\cite{elsalamony2016detection} proposed an algorithm to detect 3 classes (sickle cell, elliptocytosis and unknown shapes). After a classification of healthy/unhealthy cells, using a set of 6 geometric features and three ANNs (one for each target class), the algorithm would be used to classify unhealthy cells. The output targets were measured with regard to the solidity and circularity. 

Several works compared different classifiers~\cite{das2012quantitative,chen2014automatic,lotfi2015detection,rodrigues2016morphological}. Das et al.~\cite{das2012quantitative} proposed a method for 6-class classification considering a set of 25 features and five supervised classifiers (Multivariate logistic regression (MLR), Radial basis function neural network (RBF-NN), MLP, NB, and CART). The selected significant features were ranked based on information gain and F-value.  In this study overlapping cells were not considered. The better performance was obtained  by the MLR classifier. Chen et al.~\cite{chen2014automatic} used a set of 14 features and 3 classifiers (J48 tree, NB, and DTNB rule) to identify abnormal erythrocytes using Weka software. The proposed technique recognized normal and abnormal erythrocytes using eight directional information, irregularity and differential value from chain codes technique. The parameters for each classifier algorithm were not provided. Only abnormal erythrocytes were used as input to conduct the classification of sickle cell anemia. Lofti et al.~\cite{lotfi2015detection} used SVM, kNN and ANN to classify RBCs for iron deficiency anemia. The work used 33 determinant features which were either boundary or region descriptors, including 10 coefficients of Fourier descriptors and 7 Hu moments to classify 3 types of abnormal RBCs using SVM and kNN classifiers for each one of the 3 classes. Final classification was done by using maximum voting theory and samples classified in more than one class were eliminated. Clusters were not taking into account. 

Rodrigues et al.~\cite{rodrigues2016morphological} compared Naïve Bayes, kNN and SVM to classify a subset of isolated cells from the  erythrocytesIDB dataset~\footnote{As we can see below, this dataset is available at \url{http://erythrocytesidb.uib.es/}}. Clusters were not studied. 9 simple geometric shape features were considered and the best performance was obtained by the SVM classifier and the set of features excluding the perimeter.  

Gual-Arnau et al.~\cite{gual_arnau2015erythrocyte}, as in~\cite{lotfi2015detection,rodrigues2016morphological}, proposed a method to automatically classify isolated  erythrocytes cells as normal, sickle cells and cells with different deformation using the kNN classifier. The method used the generalized support function and some other functions related to or derived from it, to obtain a set of features related to the cells contour. Different sets of features were compared (based on the UNL-Fourier function, Fourier descriptors, Crofton’s descriptor and the CSF-ESF shape descriptors). Sharma et al.~\cite{sharma2016detection} also proposed  kNN classifier to detect three different types of distorted RBC (sickle cells, dacrocytes and elliptocytes) responsible of SCD and thalassemia, with an accuracy of 80.6\% and sensitivity of 87.6\%. Overlapping cells were not considered. 

Besides the studies related to SCD from peripheral blood smear images, there is a large number of works related to the morphological analysis of blood cells from blood microscopic images. In~\cite{rodellar2018}, we can find a detailed account about ML methods for lymphoma and leukemia cells detection from peripheral blood smears. As was pointed by the authors, the most used methods in the classification of peripheral blood were ANN, SVM and the DT. Also in~\cite{xu2017deep,zhang2018semantic} we can see how  convolutional neural networks (CNN) were used to classify RBCs on raw microscopic images. A. Kihm et al.  in \cite{khim2018classification}  try to determine the shape of RBCs in  images of microcapillary Poiseuille Flow  using CNN's. In \cite{alzubaidi2018classification} using a deep convolutional neural network as feature extractor and later an error-correcting output codes classifier for the classification task, the authors classify RBCs in three classes with an accuracy of 92,06\%. Later, the same authors in \cite{alzubaidi2020deep}  analyze a set of deep learning models in combination with different scenarios to classify red blood cells. They conclude that their Model 2 Architecture is the most appropriate for the task at hand. In addition, they used the extracted features by the proposed deep models to train a multiclass SVM classifier, obtaining high accuracies in the data sets used. We should be note that the features extracted by the deep neural network models are very difficult to interpret and are not explainable.   We can also consult a recent work on ML methods for the diagnosis of malaria by analyzing and interpreting microscopic images of thin blood smears~\cite{devi2017erythrocyte}.

%\subsection{objective}
Despite all the works consulted, most of the studies do not consider clusters of cells and cells in the border of the image, and as far as we are concerned, we have not found works that use random forest (RF), extra trees (ET) and  gradient boosting (GB) for RBCs classification from peripheral blood smear images for SCD diagnosis. This is why the main objective of our work is to propose a methodology to select the classification method and features with best performance for diagnostic support through peripheral blood smear images of RBCs, in our case samples of patients with SCD which can be generalized for other study cases.

To accomplish this aim:
\begin{itemize}
    \item We present all the features found in the literature that were used for cell classification tasks, specifically for morphology classification of blood cells. 
    \item We present classification methods reported in the literature as successful for blood cells morphology classification tasks.
    \item We determine the best parameters for all the selected classifiers to optimize the classification of sickle cells in blood smear images in a SCD case study.
    \item We determine the best performing classification methods based on the case study.
    \item We determine the most important features for the case study.
    \item We compare the best performing classification methods with other state-of-the-art algorithms.
\end{itemize}

To trust the behavior of diagnostic support systems, they must be able to explain their decisions and actions to human users~\cite{adadi2018peeking}. Interpretable ML can cover the entire model, individual components as features, and the level of training~\cite{lipton2018mythos}. As an important component of interpretable ML are the features, a key part of this work is to rank the features by their importance to guide our research towards interpretability of the proposed diagnostic support system.

This work is organized as follows: Section~\ref{sec:method} presents the applied methods. Section~\ref{sec:experimental_framework} describes the conducted experiments. Section~\ref{sec:results} presents the results and discusses these. The conclusion and future work are presented in Section~\ref{sec:conclusion}.     
    
\section{Method}
\label{sec:method}
\begin{figure}[t]
             \begin{subfigure}{0.32\textwidth}
                \centering
                \includegraphics{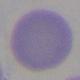}
                \caption{normal}
                \label{fig:normal}
            \end{subfigure}%
            \begin{subfigure}{0.32\textwidth}
                \centering
                \includegraphics{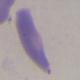}
                \caption{sickle}
                \label{fig:sickle}
            \end{subfigure}%
            \begin{subfigure}{0.32\textwidth}
                \centering
                \includegraphics{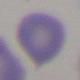}
                \caption{other}
                \label{fig:other}
            \end{subfigure}%
            \centering
            \caption{Examples of the three types of red blood cells to take into account in SCD diagnosis.}
            \label{fig:cell_types}
        \end{figure}

In this section, we propose a methodology to select the classification method and features with best performance for cell morphology analysis. Concretely, for SCD diagnostic support, there are three types of RBC to take in account based on their morphology. They can be normal (discocyte), deformed elongated (sickle cells), or cells with other deformations, as shown in Figure~\ref{fig:cell_types}.

\begin{figure}[htbp!]
\includegraphics[width=10.5cm]{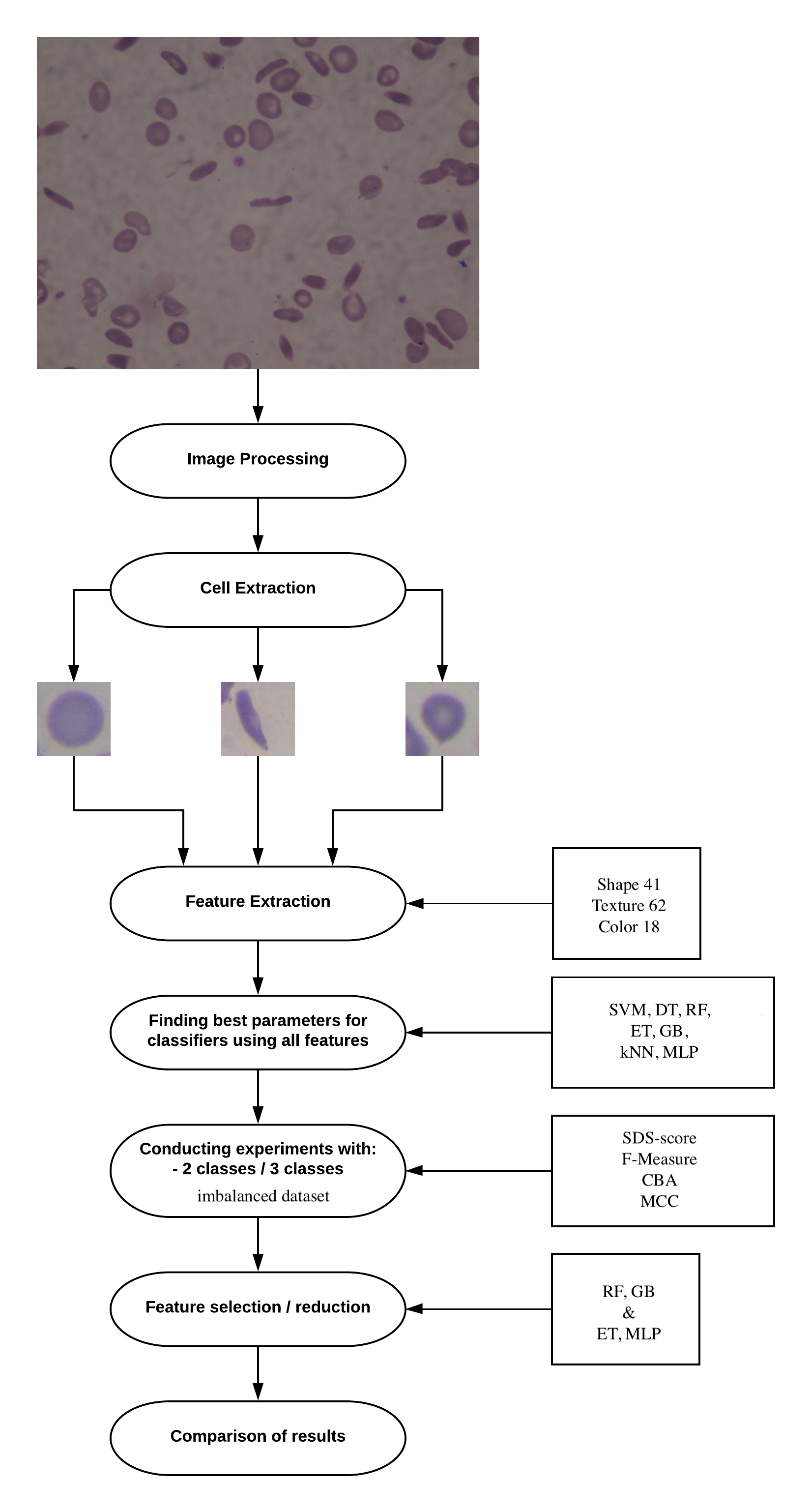}
\centering
\caption{Methodology steps of a cell classification process: image preprocessing and segmentation (cell extraction), feature extraction and feature selection/reduction, and finally classification and comparison.}
\label{fig:classification}
\end{figure}

Figure~\ref{fig:classification} shows the methodology steps based on three main phases: image pre-processing and segmentation, feature extraction, and classification.

%\begin{itemize}
%\item Image pre-processing and segmentation: To ensure high feature quality, it is important to properly process the raw data. In our work, the raw data are the microscopic blood images that need to be pre-processed to reduce noise and extract the cells to be classified. In this section, we describe all the methods we used for image pre-processing.
%\item Feature extraction: We describe the methods used to extract the features from blood cells.
%\item Classification: We describe machine learning methods to classify RBC morphology and we search for their best parameters from the resulting data in the feature extraction phase. Then, we study the characteristics the classifiers and their features. Finally, we describe the methods we used to find the best parameters for every classifier.  
%\end{itemize}

\subsection{Image pre-processing}

\begin{figure}
     \includegraphics[height=14cm]{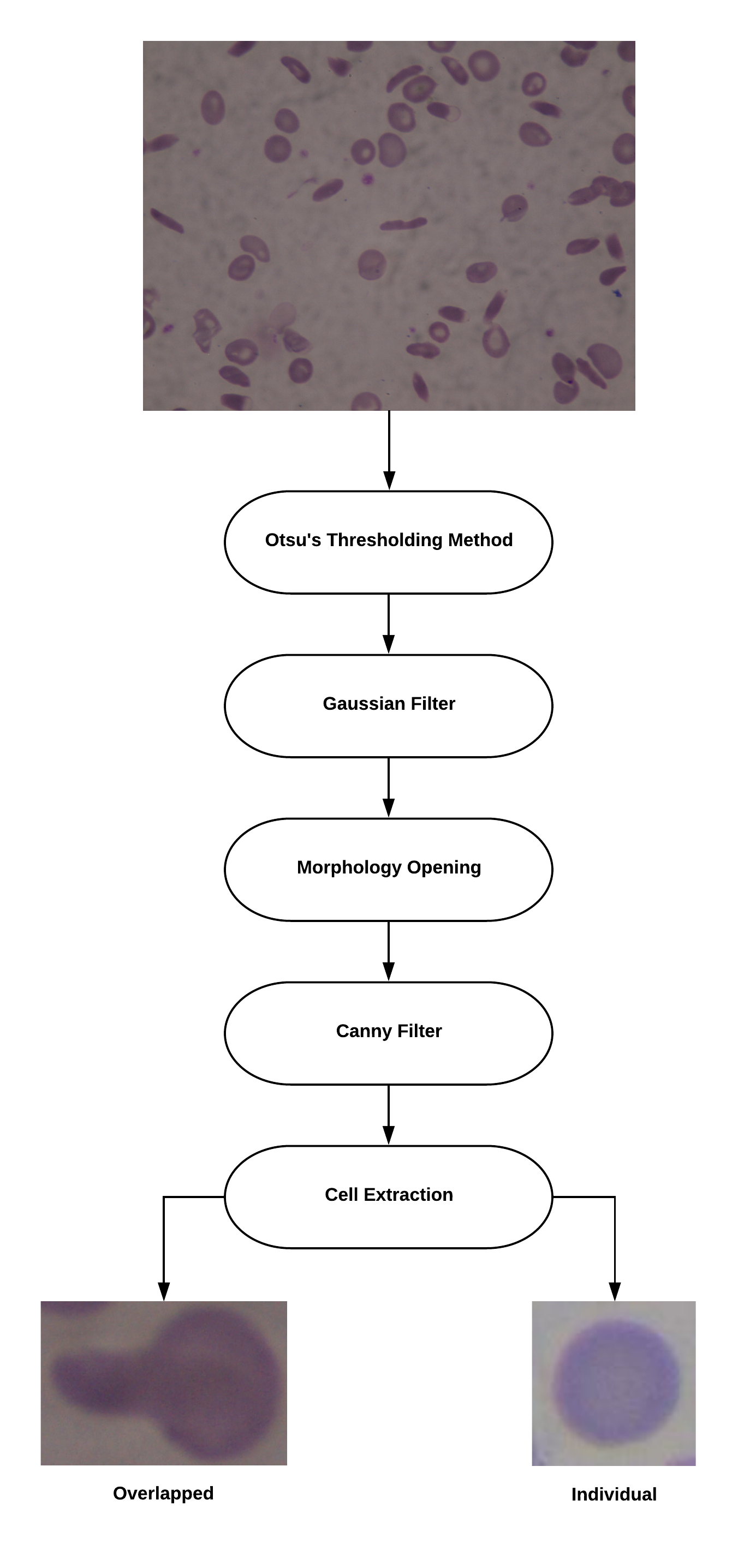} 
     \centering
    \caption{Image pre-processing and segmentation steps of the cell extraction process.}
    \label{fig:processing}
\end{figure}

The first step to be taken before feature extraction is image pre-processing. This step is needed to reduce noise and ensure the quality of the extracted features~\cite{savkare2012automatic}. Our main goal is to extract individual cells from microscopic images. An original image and the pre-processing steps taken are shown in Figure~\ref{fig:processing}.

To prepare an image for cell extraction, we used standard methods for image processing. First, we binarized the original image using Otsu's thresholding. We chose this method because of the automatic optimal threshold value calculation it provides. Second, we reduced the image noise using a Gaussian filter and we eliminated small objects using morphology opening. This method performs erosion of an image followed by its dilation. All these steps lead to the most important one, which is edge detection. Edges of the image are recognized using the Canny filter, which uses a multi-stage algorithm to detect a wide range of edges in cell images~\cite{canny1986computational}. As a result of this process, we can extract the contours of every object in the image.

A problem we encountered was that some objects represent overlapping cells, as shown in Figure~\ref{fig:processing}. In this case, an object's contour is shared by two or more cells. We solved this issue by applying a method based on ellipse adjustments and an algorithm for detecting notable points, as established by Gonz\'{a}lez-Hidalgo et al.~\cite{gonzalez2015red}. 

As a result of this step we obtained a set of Regions of Interest (ROI) where each of them contains a cell.

%%%%%%%%%%%%%%%%%%%%%%%%%%%%%%%%%%%%%%%%%%%%%%%

\subsection{Feature extraction}

Feature extraction is the first step in any classification task using ML approaches, and is shown in Figure \ref{fig:classification}. Existing literature for blood cell detection in microscopic images has proposed many features \cite{maity2017ensemble, bhowmick2013structural,yang2008survey}. We attempted to identify the three types of cells shown in Figure \ref{fig:cell_types} in our classification task. 

Regarding the identified cell types, we found the most-used features in the literature that can differentiate each type. Based on the state of the art, the extracted features belong to three categories: shape or geometry, color, and texture. We extracted forty-one shape features, eighteen color features, and sixty-two texture features from each cell for a total of 121 features to facilitate classification of the different cell types. 

\subsubsection{Shape features}
Shape features provide measures of the cells based on their geometrical properties. We developed a shape feature extraction module that extracts forty-one different geometrical features from each segmented cell. The features and their descriptions are listed in Table~\ref{table:shape_features}. Let $C$ be a closed curve enclosing a cell, and let $O$ be the area enclosed by $C$. By $|C|$ we denote the number of pixels belonging to $C$, that is the length of the boundary cell, and by $|O|$ we denote the area enclosed by $C$ (the area of the cell).  Let $\mathbf{t}_i$ be the tangent line to each point $P_i\in C$ and $d$ be the euclidean distance, then $d(\mathbf{t}_i, \mathbf{t}_j)$ measure the distance between two tangent lines to two points of $C$.  At some point, we can also denote by $\partial O $ the boundary of a region $O$.
  
\begin{landscape}
    \begin{longtable}{L{.2\textwidth} L{.45\textwidth} C{.4\textwidth} L{.5\textwidth}} \\
        \hline
        \textbf{Shape feature} & \textbf{Description} & \textbf{Equation} & \textbf{Ref} \\
        \hline
        Perimeter & Number of boundary pixels & 
        $|C|$ 
        & \cite{wheeless1994classification,markiewicz2005automatic,das2012quantitative,maity2012computer,veluchamy2012feature,bhowmick2013structural,lotfi2015detection,rodrigues2016morphological,elsalamony2016detection,xu2017deep,acharya2017,rodellar2018} \\ \hline
        Area & Number of pixels in the region inside the perimeter. & 
      $|O|$ & \cite{bacus1976image,wheeless1994classification,markiewicz2005automatic,das2012quantitative,maity2012computer,veluchamy2012feature,bhowmick2013structural,lotfi2015detection,rodrigues2016morphological,maity2017ensemble,xu2017deep,acharya2017,elsalamony2016detection,rodellar2018}\\ \hline
        Min R ($mR$) & Maximum circle radius enclosed by the cell. & & \cite{maity2012computer,chen2014automatic,maity2017ensemble} \\ \hline
        Max R ($MR$)  & Minimum circle radius that encloses the cell. & & \cite{maity2012computer,chen2014automatic,maity2017ensemble} \\ \hline
        Max feret diameter ($MFD$) & The maximum distance between parallel tangents to the boundary of the cell. & $\displaystyle \max_{P_i,P_j\in C}d(\mathbf{t}_i, \mathbf{t}_j)$ & \cite{maity2012computer,maity2017ensemble,xu2017deep} \\ \hline
        Min feret diameter ($mFD$) & The minimum distance between parallel tangents to the boundary of the cell. &  $\displaystyle \min_{P_i,P_j\in C}d(\mathbf{t}_i, \mathbf{t}_j)$ & \cite{maity2012computer,maity2017ensemble,xu2017deep} \\ \hline
        Major axis ($MA$) & The length of the major axis of the ellipse that fits $O$ by least square method. & & \cite{wheeless1994classification,das2012quantitative,markiewicz2005automatic,maity2012computer,bhowmick2013structural,lotfi2015detection,sharma2016detection,maity2017ensemble,xu2017deep}\\ \hline
        Major axis angle ($\alpha MA$) & The orientation of the major axis. & & \cite{maity2012computer,bhowmick2013structural,maity2017ensemble,xu2017deep} \\ \hline
        Minor axis ($mA$) & The length of the minor axis of the ellipse that fits $O$ by least square method. & & \cite{markiewicz2005automatic,das2012quantitative,maity2012computer,maity2012computer,bhowmick2013structural,lotfi2015detection,sharma2016detection,maity2017ensemble,xu2017deep} \\ \hline
        Convex hull perimeter & The perimeter of  convex hull $\tilde{O}$ of the shape  that enclosed the object $O$. & $|\partial \tilde{O}|$  & \cite{maity2012computer,maity2017ensemble,xu2017deep}\\ \hline
        Convex hull area & The area of convex hull. & $|\tilde{O}|$ & \cite{markiewicz2005automatic,das2012quantitative,maity2012computer,lotfi2015detection,rodrigues2016morphological,elsalamony2016detection,maity2017ensemble,xu2017deep}\\ \hline
        Aspect ratio & Ratio between major axis and minor axis. & $\frac{MA}{mA}$ & \cite{maity2012computer,deb2014noble,lotfi2015detection,sharma2016detection,xu2017deep,maity2017ensemble,acharya2017}\\ \hline
        {Eccentricity} & {Ratio between minor axis and major axis.} & $\frac{mA}{MA}$ & \cite{maity2012computer,deb2014noble,sharma2016detection,xu2017deep,maity2017ensemble,acharya2017}\\ \hline
        Circularity (also know as Form Factor, or compactness) & Measure of how closely the object approaches a mathematically perfect circle (isoperimetric ratio). More sensitive to small scale changes that increase perimeter that roundness. & $\displaystyle \frac{4 \cdot \pi \cdot  |O|}{|C|^2}$ & \cite{wheeless1994classification,asakura1996percentage,das2012quantitative,maity2012computer,veluchamy2012feature,gual_arnau2015erythrocyte,gonzalez2015red,lotfi2015detection,sharma2016detection,rodrigues2016morphological,elsalamony2016detection,maity2017ensemble,xu2017deep,acharya2017,rodellar2018}\\ \hline
        Roundness & As particle becomes circular, value approaches one. Less sensitive to perimeter changes than circularity. & $\displaystyle\frac{4 \cdot |O|}{\pi \cdot  MA^2}$ & \cite{maity2017ensemble,das2012quantitative, maity2012computer}\\ \hline
        Area equivalent diameter & The diameter of a sphere having the same projection area as particle, or diameter of the circle whose area is same as the cell area. & $\displaystyle\sqrt{\frac{4\cdot |O|}{\pi}}$ & \cite{maity2012computer,das2012quantitative,rodrigues2016morphological,lotfi2015detection,maity2017ensemble}\\ \hline
        {Perimeter equivalent Diameter} & {The diameter of a circle having the same perimeter as the projection area of the shape.} &  $\frac{|C|}{\pi}$ & \cite{das2012quantitative,maity2012computer,rodrigues2016morphological,maity2017ensemble}\\ \hline
        Equivalent Ellipse Area & Area of the ellipse that fits $O$ by least square method. & $(\pi \cdot  MA \cdot mA)/4$ & \cite{maity2012computer,maity2017ensemble,xu2017deep}\\ \hline
        Compactness & Degree to which a shape is compact with regard to perfect circle. &$\displaystyle \frac{\sqrt{\frac{4\cdot |O|}{\pi}}}{MA}$ & \cite{maity2012computer,maity2017ensemble,inproceedings}\\ \hline
        Solidity & Measures the density of an object. The ratio  of the shape area to the area of its convex hull. The proportion of pixels of convex hull that are also in the shape. & $\frac{|O|}{|\tilde{O}|}$ & \cite{das2012quantitative,maity2012computer,lotfi2015detection,elsalamony2016detection,rodrigues2016morphological,maity2017ensemble,xu2017deep}\\ \hline
        Concavity & Number of pixels that belongs to the convex hull but not to shape. A low concavity implies that the shape is really convex, but in our context this implies that the particle is well rounded.  & $|\tilde{O}| - |O|$ & \cite{markiewicz2005automatic,das2012quantitative,maity2012computer,maity2017ensemble} \cite{inproceedings}\\ \hline
        Convexity & A measure of convexity that is less sensitive to area changes. Is a measure of the particle edge roughness.  & $\frac{|\partial \tilde{O}|}{|C|}$ & \cite{wheeless1994classification,maity2012computer,maity2017ensemble}\\ \hline
        Shape & Ratio of squared perimeter to area of the object. The larger its value, the more the shape of the convexity moves away, and it can have holes. & $\frac{|C|^2}{|O|}$ & \cite{bacus1976image,wheeless1994classification,markiewicz2005automatic,das2012quantitative,veluchamy2012feature,maity2012computer,tomari2014computer,maity2017ensemble}\\ \hline
        RFactor & Ratio between a circumference of radius $\frac{MA}{2}$ and the booundary of the convex hull  & $\displaystyle \frac{|\partial \tilde{O} |}{MA\cdot \pi}$ & \cite{maity2012computer,akrimi2014classification,maity2017ensemble}\\ \hline
        Modification ratio & Ratio between the diameter of enclosed circle and the length of the major axis of the ellipse that fits $O$. & $\frac{2\cdot mR}{MA}$ & \cite{maity2012computer,akrimi2014classification,maity2017ensemble}\\ \hline
        Sphericity & Measures the degree to which a shape approaches the shape of a ``sphere''.  & $\displaystyle \frac{mR}{MR}$ & \cite{maity2012computer,akrimi2014classification,maity2017ensemble}\\ \hline
        Bounding box area (BBA) & Area of the rectangle that circumscribes the object. & $MFD \cdot mFD$ & \cite{markiewicz2005automatic,maity2012computer,maity2017ensemble}\\ \hline
        Rectangularity & Ratio of area of the object to its bounding box area. & $ \frac{|O|}{\mbox{BBA}}$ & \cite{maity2012computer,lotfi2015detection,maity2017ensemble}\\ \hline
       {Shape Factor 1} & The ratio of Feret minimun to Feret maximum & $\frac{mFD}{MFD}$ & \cite{wheeless1994classification,yang2008survey,maity2012computer,maity2017ensemble,xu2017deep}\\ \hline
        Elongation & Ratio between Max feret diameter and Min feret diameter. & $\frac{MFD}{mFD}$ & \cite{wheeless1994classification,lotfi2015detection,veluchamy2012feature,acharya2017,rodrigues2016morphological,xu2017deep}\\ \hline
        Fourier descriptors (FD1,$\ldots$, FD4) & The magnitudes of the first 4 Fourier descriptors. &  & \cite{das2012quantitative,akrimi2014classification,deb2014noble,lotfi2015detection,gual_arnau2015erythrocyte}\\ \hline
        Invariant moments (HU1,$\ldots$, HU7) & Hu moments. & \begin{tabular}{@{}l@{}l@{}l@{}l@{}l@{}l@{}l@{}l@{}l@{}l@{}l@{}l@{}l@{}l@{}l@{}l@{}l@{}l@{}}
            $I_1 = \eta_{20} + \eta_{02}$ \\ 
            $I_2 = (\eta_{20} - \eta_{02})^2 + 4\eta_{11}^2$\\ 
            $I_3 = (\eta_{30} - \eta_{12})^2 + (3\eta_{21} - \eta_{03})^2$ \\ 
            $I_4 = (\eta_{30} + \eta_{12})^2 + (\eta_{21}+\eta_{03})^2$ \\ 
            $I_5 = (\eta_{30} + 3\eta_{12})[(\eta_{30}+\eta_{12})^2 $\\ 
            $ - 3(\eta_{21} + \eta_{03})^2] + (3\eta_{21} - \eta_{03})$\\
            $(\eta_{21} + \eta_{03})[3(\eta_{30} + \eta_{12})^2 $ \\
            $ - (\eta_{21} + \eta_{03})^2]$ \\
            $I_6 = (\eta_{20}-\eta_{02})[(\eta_{30} + \eta_{12}^2) $ \\
            $ - (\eta_{21} + \eta_{03})^2]) $ \\
            $ + 4\eta_{11}(\eta_{30} + \eta_{12})(\eta_{21} + \eta_{03})$\\
            $I_7 = (3\eta_{21}-\eta_{03})(\eta_{30}+\eta_{12})$ \\
            $[(\eta_{30} + \eta_{12})^2-3(\eta_{21}+\eta_{03})^2]$ \\
            $-(\eta_{30} - 3\eta_{12})(\eta_{21} + \eta_{03})$ \\
            $[3(\eta_{30} + \eta_{12})^2 - (\eta_{21} + \eta_{03})^2]$,\\
            where $\eta_{ij} = \frac{\mu_{ij}}{\mu_{00}^{1+\frac{i+j}{2}}}$ \\
            and $\mu_{pq} = \sum_x\sum_y(x-\overline{x})^p(y - \overline{y})^qf(x,y)$\\
            where $f(x,y)$ is an image \\
            and $\overline{x}=\frac{M_{10}}{M_{00}}, \overline{y} = \frac{M_{01}}{M_{00}}$,\\
            where $M_{ij}$ are raw moments                    
        \end{tabular} & \cite{das2012quantitative,maity2012computer,maity2017ensemble,tomari2014computer,lotfi2015detection}
        \\
        \hline
        \caption{Shape and geometric features extracted from the literature  and used in this work for RBC classification.}
        \label{table:shape_features}
    \end{longtable}
\end{landscape}

\subsubsection{Color features}
    Statistical measures based on color spaces are applied to the feature extraction of a cell. The idea is to try to differentiate various cells by color~\cite{akrimi2014classification}. We used RGB, HSV, and CIEL*a*b* color spaces, as they are reported in the literature to be the most convenient for classification tasks~\cite{markiewicz2005automatic,akrimi2014classification,ongun2001feature,rodellar2018,merino2018optimizing}. Let $I$ be a channel of a color space, $N$ be the number of pixels in a region of  interest $O$ (each detected cell) in channel $I$, and $I(p)$ be the intensity value for a pixel $p$, then the channel color mean $\mu$ and its standard deviation $\sigma$ are respectively calculated as follows:
   $$
    \mu_{I,O} = \frac{1}{N}\sum_{p\in O}I(p), \quad 
    \sigma_{I,O} = \sqrt{\frac{1}{N-1}\sum_{p\in O}(I(p) - \mu)^2}.
   $$
For each detected cell and channel, we calculated the mean value and the standard deviation,  therefore we extracted eighteen color features.

\subsubsection{Texture features}

From each cell we extracted histogram and grey-level co-occurrence matrix (GLCM) features. Besides GLCM, other textures features has been used in order to classified RBCs in blood smear images, see by example  \cite{wheeless1994classification,markiewicz2005automatic,veluchamy2012feature,akrimi2014classification,merino2018optimizing}. Histogram features are first order statistical features used to represent texture of the cell's surface, we extracted two histogram features: skewness and kurtosis.
The GLCM is widely used for feature extraction in blood cell classification tasks~\cite{rezatofighi2011automatic,das2013machine,bhowmick2013structural,akrimi2014classification,rodellar2018}. The GLCM represents the relationship between neighboring pixel intensities and characterizes the texture of an image~\cite{haralick1973textural}. In our case, the GLCM was calculated for three pixel distances (1, 2, and 3) and four angles (0, $\pi/4$, $\pi/2$, and  $3\pi/4$) for a total of 12 values for each of five GLCM features listed in Table \ref{tab:texture_features}, where we defined $P_{i,j}$ as the i,j-th element of the GLCM and $N_g$ as the number of quantised grey tones.  Therefore, we extracted 60 GLCM texture features and 62 texture features in total.

\begin{center}
    \begin{table}[!ht]
    \def\arraystretch{1.5}
        \begin{tabular}{p{3cm} p{8cm} p{3cm} }
            \hline
            Texture feature & Description & Equation\\ [0.5em]
            \hline
            Skewness & Measures the degree of asymmetry. & $\frac{1}{N}\sum\limits_{i=1}^{N}(\frac{x_{i} - \mu}{\sigma})^3$ \\
            Kurtosis & Measure of whether the data are peaked or flat relative to a normal distribution. & $\frac{1}{N}\sum\limits_{i=1}^{N}(\frac{x_{i} - \mu}{\sigma})^4 - 3$ \\
            Contrast & Measures of the intensity contrast between a pixel a its neighbor pixels. Represents the amount of local gray level variation.  & $\sum\limits_{i,j=0}^{N_g-1}P_{i,j}(i-j)^2$ \\ 
            Dissimilarity & Belongs to contrast group of features and its weights are linear.  Is  a  measure  of  distance  between  pairs  of objects  in the region of interest. & $\sum\limits_{i,j=0}^{N_g-1}P_{i,j}|i-j|$ \\
            Homogeneity & Measures the smoothness (homogeneity) of the gray level distribution; it is (approximately) inversely correlated with contrast (if contrast is small, usually homogeneity is large). & $\sum\limits_{i,j=0}^{N_g-1}\frac{P_{i,j}}{1+(i-j)^2}$\\
            Energy & Measures the textural uniformity of the gray level distribution and detects disorders in textures. A smaller number of gray levels implies a higher energy. & $\sqrt{\sum\limits_{i,j=0}^{N_g-1}P_{i,j}^2}$ \\
            Correlation & Measures the linear dependency of gray levels on those of neighboring pixels. & $\sum\limits_{i,j=0}^{N_g-1}P_{i,j}\frac{(i-\mu_i)(j-\mu_j)}{\sigma_i\cdot \sigma_j}$ \\
            \hline
        \end{tabular}
        \caption{Texture features extracted from the literature and used in this work for RBC classification. The first two rows are histogram features, and the rest rows are GLCM features.}
        \label{tab:texture_features}
    \end{table}
\end{center}

\subsection{RBC classification}
In this section, we present the methods used for RBC classification. First, we explain the problems that we encountered during the data pre-processing and discuss the methods used for solving them. Second, we describe in detail the classifiers we chose to compare. Finally, we describe the process to select the best parameters for each classifier in order to compare them.
    
\subsubsection{Data pre-processing}
Before applying classification methods, we needed to analyze the data to resolve potential problems and make classification as accurate as possible~\cite{friedman2001elements}.

The first problem we encountered was that dataset values resulting from the feature extraction process had different units and ranges. This problem can affect classification results, especially regarding classifiers such as the SVM, MLP, and can also affect feature reduction methods such as principal component analysis (PCA). Non standardized features lead to 
poor performance due to faster updates of certain weights. To resolve the 
problem, we used the standard scaling method. We chose this method because it scales the feature values in way that their distribution is centered around 0 with a standard deviation of 1, and because all features are numerical values. Standard scores are calculated as follows:

\begin{equation}\label{normalization} 
z = \frac{x - \mu}{\sigma},
\end{equation}    
where $\mu$ is the mean and $\sigma$ is the standard deviation. 

\subsubsection{Classification}
\label{sec:classification}
From earlier studies in the area of blood cell classification, we recognized seven different classifiers reported to perform well at a given task. The descriptions of the classifiers are as follows: 
\begin{itemize}
    \item SVM: This statistical learning classifier, in regard to the defined classes, creates a hyper-plane to maximize the separation of data~\cite{cristianini2000introduction,zhu2017retinal}. Different studies proved robust and efficient in blood cell classification tasks~\cite{markiewicz2005automatic,ko2011cell,akrimi2014classification,amendolia2003comparative}.  

    \item Decision trees (DT): Non-parametric supervised learning method that uses simple decision rules concluded from data to predict output values~\cite{friedman2001elements}. We used the classification and regression trees (CART) algorithm. The main reason that we chose this classifier is its simplicity. Also, Maity et al.~\cite{maity2012computer} reported that DT was very successful in classifying different types of erythrocytes. Furthermore, we wanted to evaluate how this simple method compares to other, more complex algorithms.

    \item Random forest (RF): It combines a number of fully grown decision tree classifiers developed from various dataset subsamples. It uses averaging to improve the predictive accuracy and to control overfitting. We chose to compare it with other classifiers owing to the evidence that it can outperform SVM~\cite{ko2011cell}.
    
    \item Extra trees (ET): This classifier is also known as extremely 
    randomized trees~\cite{geurts2006extremely}. It is a variant of RF and it differs in that it splits nodes by choosing random cut-points and uses the whole learning sample at each step. In~\cite{Maree2007} satisfactory performance of ET was reported on an RBC dataset of isolated cells, taken in a capillary $0.05\times 0.5$ mm suspension of cells using an ACM microscope \cite{Schonfeld1989automatic}.

    \item Gradient boosting (GB): It combines weak classifiers, same as RF, and reduces the resulting error. It combines weak decision trees, which contain two leaves and have high bias and low variance. GB reduces the error by lowering the bias~\cite{friedman2001elements}.
    
    \item k-nearest neighbors (kNN): This classification method is a type of non-generalized learning. It does not create a general model, but instead stores instances of the training data. Classification is computed from a majority vote of the nearest neighbors of each point. We used it because it has performed well when classes have irregular boundaries, although it is a simple classifier~\cite{friedman2001elements}. In \cite{lotfi2015detection} 33 features were used as inputs to three machine learning classification methods: Neural Networks, SVM and kNN for RBC classification by using maximum voting. 
    
    \item Multilayer perceptron (MLP): It is the one of the most widely used artificial neural networks algorithms. It is composed of layers of input and output elements related through weighted connections. We used a MLP algorithm that learns using backpropagation. It has been reported as performing well for blood cell classification in general~\cite{ongun2001feature}, and we wanted to test it for classification of red blood cells exclusively.
\end{itemize}

\subsubsection{Hyperparameter Selection}
\label{sec:paramselection}
Once we selected the classifiers, we wanted to compare them. First we selected the best parameters for each one, then we compared the performance of each classifier. Selecting the best parameters to compare all classifiers was conducted by two methods: Randomized and Grid search.

Randomized search method randomly generates the subset of a given range of possible parameter values and evaluates the performance regarding the value of the defined metric. Grid search is an exhaustive search over every subset of the defined hyper-parameter space.

To avoid overfitting we used a 10-fold cross-validation procedure~\cite{friedman2001elements} to evaluate classifiers performance using both Randomized and Grid search. Dataset was split into training and testing sets with ratio 70/30. Training set was used for fine tuning the classifiers and the prediction was made on the testing set using the best parameters resulting from the Randomized and Grid search. %Parameter tuning process is shown in Figure~\ref{fig:tuning}.

%    \begin{figure}[htbp!]
%        \includegraphics[width=12cm]{fine_tuning.jpeg}
%        \centering
%        \caption{Parameter tuning process.}
%        \label{fig:tuning}
%\end{figure}

\section{Experimental Framework}
\label{sec:experimental_framework}
In order to validate our work, we presented a case study for microscopic images of patients with SCD. In this section we described in detail the dataset that we used. Furthermore, we explained the metrics used to compare the classifiers. Finally, we described the conducted experiment, which has three parts. First part consists of tuning classifiers to find the best parameters for given dataset with all features. In second part, we used feature reduction and selection techniques to improve the results from the first part of the experiment and also to advance towards interpretable models. Finally, we compared our results with the state of the art of RBC morphological classification. All the cells were classified regarding previously defined three classes.

All the experimentation of this article have been performed using Python 3.6. All classifiers used, belong to the Python scikit-learn library. The code that we implemented is available at \url{https://gitlab.com/miquelca32/features/}. Experiments were run on the machine with AMD Ryzen 5 3600 6-Core processor 3.60 GHz, 32 GB of RAM and Windows 10 operating system.

\subsection{Dataset}
\label{sec:dataset}
The microscopic images of the blood smears used in this work were collected from erythrocytesIDB~\cite{gonzalez2015red}, available at http://erythrocytesidb.uib.es/. The images consist in 
prepared samples of patients with sickle cell anemia classified by a specialist from ``Dr. Juan Bruno Zayas'' Hospital General in Santiago de Cuba, Cuba.

Samples were obtained from voluntary donors. The donor's thumb was pricked with a lancet and a drop of blood was collected on a sheet. The blood was spread with a coverslip at an angle of $ 45^{\mathrm{o}} $ with respect to the sheet, allowed to dry, and then fixed with a May-Gr\"unwald methanol solution. May-Gr\"unwald-Giemsa staining is a frequently used method for blood smears. 
    
The images were acquired using a Leica microscope (100$ \times $) and a Kodak EasyShare V803 camera (Kodak Retinar Aspheric All Glass Lens of 36--108 mm AF 3$ \times $ optical).

Every image was labeled by the medical expert and the specialist's criteria were used as an expert approach to validate the results of the classification methods used in this paper.

In the dataset there are a total of 2695 tagged cells (1663 circular, 700 elongated and 332 other cells). Overlapping cells were included with a total of 66 clusters ($7.1\%$ of the cells).

\subsection{Evaluation metrics}
\label{sec:evalmetrics}
To be able to compare the performance of different classifiers we used the well-known F-measure, and SDS-score~\cite{delgado2020diagnosis} as an indicator that the results provided by the method are useful for the diagnosis support of SCD. SDS-score is justified with the fact that the misclassification of the normal cells as the elongated or other cells will cause the alert to the medical specialist that the patient's condition has worsened and that the therapy should be changed. Then, it is up to the specialist to review the diagnosis and to decide whether the more drastic treatment should be prescribed. This type of error is not so serious because the treatment usually has no side effects. More dangerous scenario would be to classify deformed cells (elongated or other) as normal. In this case, the specialist could decide  that the patient is not at risk of a vaso-occlusive crisis, and the necessary treatment would not be applied. To support the diagnosis in a good way, classifiers needs to minimize the missclassification rate of elongated cells and cells with other deformations as normal cells, and the missclassification of normal cells as elongated and cells with other deformations.

We also used Class Balance Accuracy (CBA)~\cite{mosley2013} and Mathews Correlation Coefficient (MCC)~\cite{Gorodkin2004} for imbalanced classes, due to the use of overall field of view of the image, where the existing quantity of normal cells is much greater than the quantity of elongated or other deformation cells. The study of performance measures for multiple class problems is an active research field~\cite{kautz2017generic,Branco2017}. 

All of these measures are derived from the confusion matrix of true positives (TP), true negatives (TN), false positives (FP) and false negatives (FN) for all three classes.

\subsection{Experiment 1: Selection of best classifiers}
\label{sec:descripexperiment1}
The goal of the first experiment was to select the best performing classifiers on the available dataset. The best classifiers are the result of the performance comparison described in Section~\ref{sec:evalmetrics}.

First, we calculated each classifier performance using the baseline parameters defined by the Python's Scikit library and over all the set of features using 10-fold cross-validation procedure to avoid overfitting. Second, we performed a fine tuning procedure on each of them. Dataset was split to train and test set in 70/30 ratio. The fine tuning consisted of selecting the values for each parameter that ensures the best performance of each classifier regarding the average F-measure, and then regarding the SDS-score.  Fine tuning was performed using randomized search with 10-fold cross-validation on the training set. Having selected the best parameters for every classifier, we made a prediction on the testing set to obtain the final results. We compared all the results and finally, selected the best performing classifiers.

\subsection{Experiment 2: Selection of the most important features}
\label{sec:descripexperiment2}
The goal of the second experiment was to select the most important features for this classification task while preserving or even improving the performance of the chosen classifiers. The idea was to reduce the number of the features because they may cause unnecessary noise in the classification process. Furthermore, the reduction of features decreases the model's complexity and training time. Also, as features are an important component of interpretable machine learning~\cite{lipton2018mythos} we would like to rank them by their importance to analyse if this can explain the diagnosis.The feature selection was conducted in two parts.

First, we took best classifiers from previous experiment and we ranked their features by their importance. Then we trained each classifier using a wrapper method to select the best feature set, starting from the best one and iteratively increasing the number of features used, until the classifier performance was maximized. At each iteration best parameters were found performing a grid search with 10-fold cross-validation on the training set. The evaluation metrics were the F-measure and the SDS-score. The results were obtained using the best parameters to make prediction on the testing set.

Second, we tested feature reduction techniques, that is,  Principal Component Analysis (PCA) and Linear discriminant analysis (LDA).  Regarding PCA, we tested best classifiers from first experiment with different number of principal components, up to explain more than 95\% of variance. 

Finally we wanted to compare the different sets of best features we obtained. We tested if this sets could improve the results obtained in first experiment for all classifiers.
We conducted the fine tuning performing a grid search with 10-fold cross-validation on the training set of each classifier described in Section~\ref{sec:paramselection} with the different sets of features obtained from PCA, LDA and the ones obtained from best classifiers. 

\subsection{Experiment 3: Comparison with the state-of-art}
\label{sec:descripexperiment3}
The goal of the third experiment is to compare best classifiers selected from previous experiment with other state-of-art algorithms. The follow state-of-the-art on automatic counter of RBCs based  on erythrocyte shape descriptors  were considered in order to compare them with the method proposed in this paper. 

In \cite{asakura1996percentage}, Asakura determines the percentage of sickle erythrocytes cells with three different morphologies by measuring the area, perimeter, and short-axis/long-axis ratio of each cell, that is, the CSF (circular shape factor) and ESF (elliptical shape factor), that are our circularity and eccentricity, respectively (see Table \ref{table:shape_features}).

In \cite{acharya2018identification}, Acharya and Kumar presented a method for automatically classify RBCs in two classes: normal cell (circular shape) and abnormal cell. After a conversion to grey scale, the image segmentation was performed using the global Otsu threshold method. Morphological operations and modified watershed transform were applied to the image segmentation to separate the touching objects. After that, two possibilities are open: counting  the cells using the labeling algorithm and the CSF (form factor or our circularity) to identify normal cells, or counting  the cells using circular Hough transform.

In \cite{gonzalez2015red}, Gonz\'alez-Hidalgo \textit{et al.}
proposed a method for the analysis of the shape of erythrocytes in peripheral blood smear samples of sickle cell disease, which uses ellipse adjustments and an algorithm for detecting notable points. They  applied a set of constraints that allowed to eliminate significant image preprocessing steps proposed in previous studies. They classify cells between circular and elongated using the ESF (our eccentricity).

\begin{center}
\begin{table}[t!]
   \footnotesize
        \begin{tabular}{L{1.2cm}L{7cm}L{7cm}}
            \hline
            Classifier & SDS-score Parameters  & F-measure Parameters \\
            \hline

            SVM & kernel='rbf', C=10, gamma=0.01 & kernel='rbf', C=10, gamma=0.01\\ 
            \hline
            DT & max\_depth=None, max\_features=None, min\_samples\_leaf=10, min\_samples\_split=10, splitter='best' & max\_depth=None, max\_features=None, min\_samples\_leaf=10, min\_samples\_split=10, splitter='best'\\
            \hline
            RF &  n\_estimators=300, min\_samples\_leaf=2, min\_samples\_split=2, max\_depth=None, max\_features='sqrt', bootstrap=True & n\_estimators=300, min\_samples\_leaf=2, min\_samples\_split=2, max\_depth=None, max\_features='sqrt', bootstrap=True\\
            \hline
            ET & n\_estimators=60, max\_depth=None, max\_features='auto', min\_samples\_leaf=1, min\_samples\_split=10, bootstrap=True & n\_estimators=60, max\_depth=None, max\_features='auto', min\_samples\_leaf=1, min\_samples\_split=10, bootstrap=True\\
            \hline
            GB & sub\_sample=1, min\_child\_weight=1, max\_depth=10, max\_delta\_step=10 & sub\_sample=1, min\_child\_weight=1, max\_depth=10, max\_delta\_step=20\\
            \hline
            kNN & n\_neighbors=10, weights='distance', p=1, algorithm='brute' & n\_neighbors=9, weights='distance', p=1, algorithm='brute'\\
            \hline
            MLP & solver='sgd', hidden\_layer\_sizes=(10, 3), activation='identity' & solver='sgd', hidden\_layer\_sizes=(10, 3), activation='identity'\\
           \hline
        \end{tabular}
        \caption{Best parameters selected by fine tuning  that maximize the  SDS-score and  F-measure   performance measures,. respectively.}
        \label{tab:best_params}
    \end{table}
\end{center}

\section{Results and discussion}
\label{sec:results}
The aims of the experiments are to determine the most appropriate classifier to successfully support the diagnose of the SCD, to find the set of most important features and to compare our results with other state-of-the-art algorithms.

\subsection{Experiment 1}

%First experiment was designed to select best classifiers for the cell classification task. All classifiers were trained using the complete set of features and we performed a randomized search over the parameters of each classifier in order to maximize the SDS-score and F-measure metrics. 
Recall that, as was pointed in Section \ref{sec:descripexperiment1}, this  experiment was designed to select best classifiers for the cell classification task. Table~\ref{tab:best_params} shows best parameters for each classifier regarding the used evaluation measures. 

Table \ref{tab:best_experiment_1} depicts the comparison of classifiers regarding F-measure and SDS-score. From the obtained results we can observe that GB, RF, DT, ET and MLP have similar performance, although the best ones are RF and GB. Regarding running times of randomized search over the same dataset, we excluded GB. It took 132 minutes for GB, 11 minutes for RF, 7 minutes for ET and only 50s for DT to run. Therefore, we chose RF as the best resulting classifier and DT as second best and fastest one to proceed with the experiments (taking account SDS-score results because indicates how useful are the results provided by the method for support the diagnosis  of SCD). Moreover, \ref{appendix-third} shows the running times of the experimentation, where running time of GB, for one iteration of Grid search and for training, takes much longer than the rest of the classifiers. The benefits obtained by classification using GB is not compensated by the time needed to train this classifier and perform the fine-tuning process. Confusion matrices obtained from best classifiers as a result of this process are depicted in Table~\ref{confusion_table} (\ref{appendix-first} shows confusion matrices for each classifier).

\begin{center}
\begin{table}[ht!]
\begin{tabular}{lccccccc}
\hline
             & SVM   & DT    & RF    & ET    & GB    & kNN   & MLP   \\ \hline
\textbf{Baseline F-measure}  & 85.97\% & 86.49\% & 89.91\% & 89.54\% & 89.75\% & 85.35\% & 85.39\% \\
\textbf{Fine tuning F-measure} & 91.50\% & 91.62\% & \textbf{93.11\%} & 92.91\% & \textbf{93.40\%} & 88.93\% & 92.23\% \\
\textbf{Baseline SDS-score} & 87.24\% & 90.02\% & 91.61\% & 89.76\% & 90.91\% & 87.31\% & 87.68\% \\
\textbf{Fine tuning SDS-score} & 93.45\% & 94.44\% & \textbf{94.56\%} & 94.31\% & \textbf{94.56\%} & 90.98\% & 94.19\% \\ \hline
\end{tabular}
\caption{Performance of the different classifiers for the F-measure and SDS-score performance measures in Experiment 1. We show the performance measures for the baseline experiment and the fine tuning parameters experiment.}
\label{tab:best_experiment_1}
\end{table}
\end{center}

\begin{center}
\begin{table}[ht!]

%\resizebox{\textwidth}{!}{%
\begin{tabular}{@{}rrrrrrrrrrrrrr@{}}
\cmidrule(r){1-4} \cmidrule(lr){6-9} \cmidrule(l){11-14}
\begin{tabular}[c]{@{}l@{}}RF\\ baseline\end{tabular} & c & e & o &  & \begin{tabular}[c]{@{}l@{}}RF\\ max F-measure\end{tabular} & c & e & o &  & \begin{tabular}[c]{@{}l@{}}RF\\ max SDS-score\end{tabular} & c & e & o \\ 
\cmidrule(r){1-4} \cmidrule(lr){6-9} \cmidrule(l){11-14} 
c & 1580 & 38 & 45&  & c & 490 & 7 & 2& & c & 490 & 7 & 2  \\
e & 52 & 636 & 12 &  & e & 9 & 197 & 4 & & e & 9 & 197 & 4 \\
o & 91 & 27 & 214 &  & o & 26 & 6 & 68 & & o & 26 & 6 & 68 \\ \cmidrule(r){1-4} \cmidrule(lr){6-9} \cmidrule(l){11-14}
\end{tabular}%
%}
%\caption{Decision tree confusion matrices comparison}
\label{confusion_gb_1}

%\resizebox{\textwidth}{!}{%
\begin{tabular}{@{}rrrrrrrrrrrrrr@{}}
\cmidrule(r){1-4} \cmidrule(lr){6-9} \cmidrule(l){11-14}
\begin{tabular}[c]{@{}l@{}}{DT}\\ baseline\end{tabular} & c & e & o &  & \begin{tabular}[c]{@{}l@{}}DT\\ max F-measure\end{tabular} & c & e & o &  & \begin{tabular}[c]{@{}l@{}}DT\\ max SDS-score\end{tabular} & c & e & o \\ \cmidrule(r){1-4} \cmidrule(lr){6-9} \cmidrule(l){11-14} 
c & 1535 & 47 & 81 &  & c & 481 & 8 & 10 &  & c & 481 & 8 & 10 \\
e & 69 & 564 & 67 &  & e & 6 & 194 & 10 &  & e & 6 & 194 & 10 \\
o & 72 & 32 & 228 &  & o & 21 & 12 & 67 &  & o & 21 & 12 & 67 \\ \cmidrule(r){1-4} \cmidrule(lr){6-9} \cmidrule(l){11-14} 
\end{tabular}%
%}
%\caption{Random forest confusion matrices comparison}
\label{confusion_rf_1}

\caption{Confusion matrices for best classifiers selected in the baseline experiment (left) and fine tuning to maximizing F-measure and SDS-score (center and right).}
\label{confusion_table}
\end{table}
\end{center}

\subsection{Experiment 2}
%The goal of this second experiment was to find the set of most important features while preserving or even improving the performance of all classifiers.

First part of the experiment consisted on using best classifiers from previous experiment to check if there were a subset of features that improve the previous obtained results. We performed an incremental search over the ordered set of all features, for RF and GB, based on its importance. We ranked the RF features based on its mean decrease impurity and we ranked the GB features based on the average of the impurity gain across each of the classifiers. For each subset of ordered features we performed a grid search of the best parameters in order to maximize the SDS-score and the F-measure. Second column of Table~\ref{tab:best_experiment_2}  shows that we improved the results obtained in the first experiment, we can observe that dimensionality reduction gives us an improvement over results of first experiment. RF result is maximized (SDS-score 95.06\%) with the 15 most important features: Aspect ratio, Elongation, R factor, HU1, Roundness, HU2, Minor axis, Shape, FD1, Circularity, Min feret, HU7, HU3, Compactness, and Skewness. Although DT result is maximized (SDS-score 94.44\%) with all the features, we considered most important features to be the first 20, when the result was over 94\% (SDS-score 94.07\%) and F-measure maximized with 92.37\%: R factor, Aspect ratio, Area equivalent diameter, Skewness, Roundness, Minor axis, Correlation1, Homogeneity12, Max R, Major axis, HU5, Min feret, Energy9, Homogeneity1, Solidity, Kurtosis, Blue mean, HU3, Contrast9, L mean. 

Second part of this experiment was to reduce the feature space using two dimensionality reduction methods, PCA and LDA over the whole set of features. We performed a search on the PCA dimensional space in order to maximize the classification measures up to explain the 95.01\% of the variance, that is selecting the first 16 components. For LDA we used 2 components explaining $100\%$ of variance. First component explained $83.97\%$ of variance. For each subset of ordered principal components we performed a grid search on RF and DT. In Table~\ref{tab:best_experiment_2}, we can state that LDA dimensional reduction outperforms PCA, {but the feature selection outperforms both PCA and LDA. More precisely, in our case study reducing features using PCA and LDA worsened the results of the first experiment.

\begin{table}[ht!]
\begin{tabular}{lcccccc}
\hline
\textbf{}       &  \thead{First \\ experiment} &  \thead{Feature \\ selection} &   PCA     & LDA     \\ \hline
\textbf{RF F-measure}  & 93.11\%          & \textbf{93.36}\%                         & 90.06\%              & 91.55\% \\
\textbf{RF SDS-score} & 94.56\%          & \textbf{95.05}\%                         & 92.34\%                 & 93.20\% \\
\textbf{DT F-measure}  & 91.62\%          & 92.37\%                          & 87.02\%               & 91.49\% \\
\textbf{DT SDS-score} & 94.44\%          & 94.07\%                         & 90.48\%                & 93.20\% \\ \hline
\end{tabular}
\caption{Comparison of the performance measures obtained using fine tuning in Experiment 1, with the obtained feature selection, PCA and LDA dimensionality reduction. For RF and DT machine learning methods we maximized the SDS-score and the F-measure. }
\label{tab:best_experiment_2}
\end{table}

Table~\ref{tab:best_experiment_lda} depicts the comparison between results from the first experiment and classification with 15 most important features regarding RF on all the classifiers. Results from DT are also obtained using RF's most important features, as they are better than using DT's most important features as showed in Table~\ref{tab:best_experiment_2}. We can observe that feature selection regarding RF and the posterior fine tuning of the parameters of each classifier gives the best performance for the SDS-score and F-measure. We can note that method with best results is GB for both measures but the results for RF are second best, for that reason these are the two selected machine leaning methods for the third experiment. \ref{appendix-second} shows confusion matrices for each classifier.

\begin{table}[ht!]
\begin{tabular}{lccccccc}
\hline
                          & SVM                         & DT                          & RF                          & ET                          & GB                          & kNN                         & MLP                                  \\ \hline
\textbf{Baseline SDS-score}     & 87.24\%                 & 90.02\%                 & 91.61\%                 & 89.76\%                 & 90.91\%                 & 87.31\%                 & 87.68\%                          \\
\textbf{Important features SDS-score}          & 94.68\%                     & 94.68\%            & \textbf{95.05\%}                     & 94.31\%                     & \textbf{95.18\%}                    & 93.57\%                     & 94.81\%                              \\
 \hline
\textbf{Baseline F-measure}      & {85.97\%} & {86.49\%} & {89.91\%} & {89.54\%} & {89.75\%} & {85.35\%} & {85.39\%}          \\
\textbf{Important features F-measure}           & 93.15\% & 92.55\% & \textbf{93.36\%} & 92.88\% & \textbf{93.50\%} & 91.77\% & 93.14\% \\
 \hline
\end{tabular}
\caption{Comparison of the performance measures between the obtained in baseline, Experiment 1, and the obtained by the different machine learning methods using feature selection.}
\label{tab:best_experiment_lda}
\end{table}

Regarding interpretability, we can observe in Table~\ref{tab:best_experiment_lda} that metrics for classifiers are not so different. Their ranges are from $93.57\%$ to $95.18\%$ for the SDS-Score, and from $91.77\%$ to $93.50\%$ for the F-measure. DT and kNN are transparent models~\cite{arrieta2020explainable}, that means that the model by itself is understandable. Moreover, we ranked the features as a first step towards post-hoc interpretability~\cite{arrieta2020explainable} to target models that are not readily interpretable by design by resorting to diverse means to enhance their interpretability such as feature relevance. As can be observed, in both machine learning methods, DT and RF, only shape/geometric and texture features were selected and leaving it aside color features.  This makes sense since we are classifying shapes according to their morphology, and color has no influence on it. Feature relevance explanation techniques aim to describe the functioning of an opaque model by ranking relevance or importance each feature has in the prediction output by the model to be interpret. 

\begin{center}
\begin{table}[ht!]
\begin{tabular}{@{}lcccc@{}}
\toprule
    & GB      & RF     & \cite{asakura1996percentage} 
    \\ \midrule
SDS-score & \textbf{95.18\%} & 95.05\% & 61.80\% 
\\
F-measure  & \textbf{93.50\%} & 93.36\% & 45,33\%  
\\
CBA & \textbf{88.39\%} & 88.06\% & 37.48\%
\\
MCC & \textbf{88.43\%} & 88.20\% & 35.43\%
\\ \hline

\end{tabular}

\caption{Comparison of F-measure and SDS-score values obtained by GB, RF and Asakura~\cite{asakura1996percentage} methods for the 3 classes classification problem.}
\label{table:three_classes}
\end{table}
\end{center}

\subsection{Experiment 3}
 
%In this experiment we wanted to compare our results with other  state-of-the-art algorithms, we divided this comparison in two parts. 
As was stated in Section \ref{sec:descripexperiment3}, the aim of this experiment is to compare our results with other  state-of-the-art algorithms. First, we compared with the method described  by Asakura \textit{et al.} 
in~\cite{asakura1996percentage} as they classify eritrocites in 3 classes.
Table \ref{table:three_classes} shows that our two best classifiers with 
F-measure and SDS-score (GB and RF) outperform results from 
Asakura~\cite{asakura1996percentage}. Table \ref{confusion_3_classes} shows the 
confusion matrices used to obtain the metrics of this part of the 
experiment.

The second part of this validation experiment was to compare with other methods that classify cells in two classes: circular and elongated. In particular we compare with the work of Acharya \textit{et al.}~\cite{acharya2018identification} and González-Hidalgo \textit{et al.}~\cite{gonzalez2015red}. Table \ref{table:two_classes}  shows that our two best classifiers with F-measure and SDS-score (GB and RF)  outperform results from Acharya~\cite{acharya2018identification} and González-Hidalgo~\cite{gonzalez2015red}. Table  \ref{confusion_2_classes} shows the 
confusion matrices used to obtain the metrics of this part of the 
experiment.

\begin{center}
\begin{table}[ht!]
%\resizebox{\textwidth}{!}{%
\begin{tabular}{@{}lrrrrrrrrrrrrr@{}}
\cmidrule(r){1-4} \cmidrule(r){6-9} \cmidrule(r){11-14}
\begin{tabular}[c]{@{}l@{}}GB\end{tabular} & c & e & o &  & \begin{tabular}[c]{@{}l@{}}RF\end{tabular} & c & e & o & & \begin{tabular}[c]{@{}l@{}} \cite{asakura1996percentage} \end{tabular} & c & e & o \\
\cmidrule(r){1-4} \cmidrule(r){6-9}  \cmidrule(r){11-14} 
c & 488 & 6 & 5 &  & c & 488 & 7 & 4 &  & c & 199 & 259 & 41 \\
e & 8 & 194 & 8 &  & e & 7 & 194 & 9 &  & e & 0 & 102 & 108  \\
o & 20 & 5 & 75 &  & o & 22 & 4 & 74 &  & o & 9 & 67 & 24\\ \cmidrule(r){1-4} \cmidrule(r){6-9} \cmidrule(r){11-14} 
\end{tabular}%

%\caption{Gradient booster confusion matrices comparison}

\caption{Confusion matrices for best classifiers (GB and RF) and that   obtained by Asakura~\cite{asakura1996percentage}.}
\label{confusion_3_classes}
\end{table}
\end{center}

\begin{center}
\begin{table}[ht!]
\begin{tabular}{@{}lcccc@{}}
\toprule
    & GB    & RF    & \cite{acharya2018identification} & \cite{gonzalez2015red} 
    \\ \midrule
SDS-score & \textbf{94.68\%}  & 94.44\% & 78.49\%        & 49.32\%          
\\
F-measure & \textbf{94.67\%} & 94.42\%  & 78.76\%        & 48.97\%          
\\
CBA & \textbf{93.98\%} & 93.66\% & 81.16\% & 52.81\%
\\
MCC & \textbf{88.72\%} & 88.19\% & 60.80\% & 5.70\%
\\
\hline

\end{tabular}

\caption{Comparison of F-measure and SDS-score values obtained by GB, RF, Acharya~\cite{acharya2018identification} and González-Hidalgo~\cite{gonzalez2015red} methods  for the 2 classes classification problem.} \label{table:two_classes}
\end{table}
\end{center}

\begin{center}
\begin{table}[ht!]
%\resizebox{\textwidth}{!}{%
\begin{tabular}{@{}lrrrrrrrrrrr@{}}
\cmidrule(r){1-3} \cmidrule(lr){4-6} \cmidrule(lr){7-9} \cmidrule(l){10-12} 
\begin{tabular}[c]{@{}l@{}}GB\\  SDS-score\end{tabular} & c & e &   \begin{tabular}[c]{@{}l@{}}RF\\  SDS-score\end{tabular} & c & e &  
\begin{tabular}[c]{@{}l@{}} \cite{gonzalez2015red} \end{tabular} & c & e & 
\begin{tabular}[c]{@{}l@{}} \cite{acharya2018identification} \end{tabular} & c & e  \\
\cmidrule(r){1-3} \cmidrule(lr){4-6}  \cmidrule(lr){7-9} \cmidrule(l){10-12} 
c & 484 & 15 &   c & 484 & 15 &    c & 189 & 310 &   c & 348 & 151 \\
e & 28 & 282 &   e & 30 & 280 &     e & 100 & 210  &   e & 23 & 287 \\
 \cmidrule(r){1-3} \cmidrule(lr){4-6} \cmidrule(lr){7-9} \cmidrule(l){10-12} 
\end{tabular}%
%}
%\caption{Confusion matrices comparison}
\label{confusion_comparison}

\caption{Confusion matrices for GB, RF and results from Acharya~\cite{acharya2018identification} and González-Hidalgo~\cite{gonzalez2015red}.}
\label{confusion_2_classes}
\end{table}
\end{center}

Regarding interpretability, we can observe in Table~\ref{tab:best_experiment_lda} that transparent models~\cite{arrieta2020explainable}, DT and kNN, obtained better results than~\cite{asakura1996percentage,gonzalez2015red,acharya2018identification}.

\section{Conclusion and future work}
\label{sec:conclusion}

 We presented a general methodology to select the classification method and features with best performance for diagnostic support through peripheral blood smear images of red blood cells, in our case samples of patients with sickle-cell disease diagnosis which can be generalized for other study cases. 
 
We collected and implemented features and classifiers that previous literature validated for the study case or similar cases. Then, we explained how to select the best parameters for each classifier. We included as results the best parameters for each classifier, the implemented code \url{https://gitlab.com/miquelca32/features/} and the confusion matrices with the raw data to allow researchers to more easily compute other metrics. The  dataset we used is available at \url{http://erythrocytesidb.uib.es/}. For the sake of scientific progress, it would be beneficial if authors published their raw data, code and the image datasets that they used.
 
Next, we defined how to select the most important features for classification in order to reduce their total number to decrease the complexity and training time. Also, feature selection allowed us to give more information about the functioning of opaque models (non interpretable) by feature ranking in the model's prediction output.
 
We compared the best performing classification methods with the state-of-the-art methods in order to validate that the proposed methodology allows to select the most appropriate ML model for a case study, in this paper microscopic images of patients with sickle cell anemia. More precisely, we obtained better results than the state-of-the-art methods with all the models used in this paper, even with interpretable model classifiers. That is a good result for health environments where there is a need to trust and audit diagnostic support systems. 
 
As a further work, we want to research on diagnosis support of other diseases through cell morphology analysis from microscopy images and advance towards more interpretable approaches. We also are interested in research on ensemble voting, subset of features for optimizing classification, and feature selection based on PCA and LDA for sickle-cell disease diagnosis support and its interpretability.

\section*{Acknowledgements}

We acknowledge the Spanish Government for its support by the Project EXPLainable Artificial INtelligence systems for health and well-beING (EXPLAINING) (PID2019-104829RA-I00 / AEI / 10.13039/501100011033), the Spanish Grant TIN2016-75404-P, AEI/FEDER, UE, and also the Spanish Grant TIN2016-81143-R, AEI/FEDER, UE. We also acknowledge the Govern de les Illes Balears for its support to the project PROCOE/2/2017. Nataša Petrović also benefited from the fellowship Euroweb$+$Program (ERASMUS-MUNDUS).

\iffalse
We acknowledge the Spanish Government, the Ministerio de Economía, Industria y Competitividad (MINECO), the Agencia Estatal de Investigación (AEI), and the European Regional Development Funds (ERDF) for their support to the projects TIN2016-81143-R (MINECO/AEI/ERDF, EU) and TIN2016-75404-P (MINECO/AEI/ERDF, EU). We also acknowledge the Govern de les Illes Balears for its support to the project PROCOE/2/2017. Nataša Petrović also benefited from the fellowship Euroweb$+$Program (ERASMUS-MUNDUS).
\fi

\section*{References}
\bibliographystyle{elsarticle-num}
\bibliography{bibliography} 

\newpage 
\appendix
\section{First experiment: confusion matrices}

\label{appendix-first}

\begin{longtable}{c}

%\begin{table*}[ht!]
%\resizebox{\textwidth}{!}{%
\begin{tabular}{@{}lrrrrrrrrrrrrr@{}}
\cmidrule(r){1-4} \cmidrule(lr){6-9} \cmidrule(l){11-14}
\begin{tabular}[c]{@{}l@{}}SVM\\ baseline\end{tabular} & c & e & o &  & \begin{tabular}[c]{@{}l@{}}SVM \\ max F-measure\end{tabular} & c & e & o &  & \begin{tabular}[c]{@{}l@{}}SVM\\ max SDS-score\end{tabular} & c & e & o \\ \cmidrule(r){1-4} \cmidrule(lr){6-9} \cmidrule(l){11-14} 
c & 1496 & 44 & 123 &  & c & 1559 & 50 & 54 &  & c & 1559 & 50 & 54 \\
e & 83 & 596 & 21 &  & e & 81 & 615 & 4 &  & e & 81 & 615 & 4 \\
o & 94 & 19 & 219 &  & o & 111 & 31 & 190 &  & o & 111 & 31 & 190 \\ \cmidrule(r){1-4} \cmidrule(lr){6-9} \cmidrule(l){11-14} 
\end{tabular}%
%}
%\caption{SVM confusion matrices comparison}
%\label{confusion_svm}

\\

%\resizebox{\textwidth}{!}{%
\begin{tabular}{@{}lrrrrrrrrrrrrr@{}}
\cmidrule(r){1-4} \cmidrule(lr){6-9} \cmidrule(l){11-14}
\begin{tabular}[c]{@{}l@{}}DT\\ baseline\end{tabular} & c & e & o &  & \begin{tabular}[c]{@{}l@{}}DT\\ max F-measure\end{tabular} & c & e & o &  & \begin{tabular}[c]{@{}l@{}}DT\\ max SDS-score\end{tabular} & c & e & o \\ 
\cmidrule(r){1-4} \cmidrule(lr){6-9} \cmidrule(l){11-14} 
c & 1535 & 47 & 81 &  & c & 1552 & 40 & 71& & c & 1552 & 40 & 71  \\
e & 69 & 564 & 67 &  & e & 51 & 620 & 29 & & e & 51 & 620 & 29 \\
o & 72 & 32 & 228 &  & o & 82 & 32 & 218 & & o & 82 & 32 & 218 \\ \cmidrule(r){1-4} \cmidrule(lr){6-9} \cmidrule(l){11-14}
\end{tabular}%
%}
%\caption{Gradient booster confusion matrices comparison}
%\label{confusion_gb}

\\

%\resizebox{\textwidth}{!}{%
\begin{tabular}{@{}lrrrrrrrrrrrrr@{}}
\cmidrule(r){1-4} \cmidrule(lr){6-9} \cmidrule(l){11-14}
\begin{tabular}[c]{@{}l@{}}RF\\ baseline\end{tabular} & c & e & o &  & \begin{tabular}[c]{@{}l@{}}RF\\ max F-measure\end{tabular} & c & e & o &  & \begin{tabular}[c]{@{}l@{}}RF\\ max SDS-score\end{tabular} & c & e & o \\ \cmidrule(r){1-4} \cmidrule(lr){6-9} \cmidrule(l){11-14} 
c & 1580 & 38 & 45 &  & c & 1596 & 40 & 27 &  & c & 1597 & 41 & 25 \\
e & 52 & 636 & 12 &  & e & 39 & 657 & 4 &  & e & 40 & 654 & 6 \\
o & 91 & 27 & 214 &  & o & 103 & 33 & 196 &  & o & 103 & 32 & 197 \\ \cmidrule(r){1-4} \cmidrule(lr){6-9} \cmidrule(l){11-14} 
\end{tabular}%
%}
%\caption{Random forest confusion matrices comparison}
%\label{confusion_rf}

\\

%\resizebox{\textwidth}{!}{%
\begin{tabular}{@{}lrrrrrrrrrrrrr@{}}
\cmidrule(r){1-4} \cmidrule(lr){6-9} \cmidrule(l){11-14}
\begin{tabular}[c]{@{}l@{}}ET\\ baseline\end{tabular} & c & e & o &  & \begin{tabular}[c]{@{}l@{}}ET\\ max F-measure\end{tabular} & c & e & o &  & \begin{tabular}[c]{@{}l@{}}ET\\ max SDS-score\end{tabular} & c & e & o \\ \cmidrule(r){1-4} \cmidrule(lr){6-9} \cmidrule(l){11-14} 
c & 1556 & 45 & 62 &  & c & 1569 & 47 & 47&  & c & 1569 & 47 & 47 \\
e & 72 & 620 & 8 &  & e & 54 & 637 & 9  &  & e & 54 & 637 & 9\\
o & 97 & 29 & 206 &  & o & 101 & 32 & 199 &  & o & 101 & 32 & 199\\ \cmidrule(r){1-4} \cmidrule(lr){6-9} \cmidrule(l){11-14}  
\end{tabular}%
%}
%\caption{Extra trees confusion matrices comparison}
%\label{confusion_et}

\\

%\resizebox{\textwidth}{!}{%
\begin{tabular}{@{}lrrrrrrrrrrrrr@{}}
\cmidrule(r){1-4} \cmidrule(lr){6-9} \cmidrule(l){11-14}
\begin{tabular}[c]{@{}l@{}}GB\\ baseline\end{tabular} & c & e & o &  & \begin{tabular}[c]{@{}l@{}}GB\\ max F-measure\end{tabular} & c & e & o &  & \begin{tabular}[c]{@{}l@{}}GB\\ max SDS-score\end{tabular} & c & e & o \\ \cmidrule(r){1-4} \cmidrule(lr){6-9} \cmidrule(l){11-14} 
c & 1569 & 41 & 53 &  & c & 1565 & 45 & 53 &  & c & 1569 & 41 & 53 \\
e & 60 & 628 & 12 &  & e & 56 & 631 & 13 &  & e & 59 & 628 & 13  \\
o & 91 & 26 & 215 &  & o & 82 & 24 & 226 &  & o & 78 & 24 & 230 \\ 
\cmidrule(r){1-4} \cmidrule(lr){6-9} \cmidrule(l){11-14} 
\end{tabular}%
%}
%\caption{Decision tree confusion matrices comparison}
%\label{confusion_dt}
%\end{table*}

\\

%\begin{table}[ht!]
%\resizebox{\textwidth}{!}{%
\begin{tabular}{@{}lrrrrrrrrrrrrr@{}}
\cmidrule(r){1-4} \cmidrule(lr){6-9} \cmidrule(l){11-14}
\begin{tabular}[c]{@{}l@{}}kNN\\ baseline\end{tabular} & c & e & o &  & \begin{tabular}[c]{@{}l@{}}kNN\\ max F-measure\end{tabular} & c & e & o &  & \begin{tabular}[c]{@{}l@{}}kNN\\ max SDS-score\end{tabular} & c & e & o \\ \cmidrule(r){1-4} \cmidrule(lr){6-9} \cmidrule(l){11-14} 
c & 1542 & 39 & 82 &  & c & 1551 & 43 & 69  &  & c & 1551 & 43 & 69 \\
e & 109 & 577 & 14 &  & e & 111 & 579 & 10 &  & e & 111 & 579 & 10 \\
o & 112 & 31 & 189 &  & o & 112 & 29 & 191 &  & o & 112 & 29 & 191 \\
\cmidrule(r){1-4} \cmidrule(lr){6-9} \cmidrule(l){11-14} 
\end{tabular}%
%}
%\caption{kNN confusion matrices comparison}
%\label{confusion_knn}

\\

%\resizebox{\textwidth}{!}{%
\begin{tabular}{@{}lrrrrrrrrrrrrr@{}}
\cmidrule(r){1-4} \cmidrule(lr){6-9} \cmidrule(l){11-14}
\begin{tabular}[c]{@{}l@{}}MLP\\ baseline\end{tabular} & c & e & o &  & \begin{tabular}[c]{@{}l@{}}MLP\\ max F-measure\end{tabular} & c & e & o &  & \begin{tabular}[c]{@{}l@{}}MLP\\ max SDS-score\end{tabular} & c & e & o \\ \cmidrule(r){1-4} \cmidrule(lr){6-9} \cmidrule(l){11-14} 
c & 1485 & 48 & 130 &  & c & 1515 & 45 & 103  &  & c & 1539 & 45 & 79 \\
e & 62 & 598 & 40 &  & e & 66 & 618 & 16 &  & e & 59 & 598 & 43 \\
o & 92 & 29 & 211 &  & o & 75 & 25 & 232 &  & o & 82 & 30 & 220 \\ \cmidrule(r){1-4} \cmidrule(lr){6-9} \cmidrule(l){11-14} 

\end{tabular}%
%}
\\
\caption{Confusion matrices for the first experiment (raw data Table~\ref{tab:best_experiment_1}).}
\label{confusion_table_resum}
\end{longtable}

\newpage
\section{Second experiment: confusion matrices}
\label{appendix-second}

\begin{longtable}{c}

\begin{tabular}{@{}lrrrrrrrrr@{}}
\cmidrule(r){1-4} \cmidrule(lr){5-8} 
\begin{tabular}[c]{@{}l@{}}RF 15 features\\ SDS-score\end{tabular} & c & e & o &   \begin{tabular}[c]{@{}l@{}}RF 15 features\\ F-measure\end{tabular} & c & e & o &\\ \cmidrule(r){1-4} \cmidrule(lr){5-8} 
c & 488 & 7 & 4 &   c & 489 & 7 & 3 &    \\
e & 7 & 194 & 9 &   e & 8 & 194 & 8 & \\
o & 22 & 4 & 74 &   o & 22 & 3 & 75 &  \\ \cmidrule(r){1-4} \cmidrule(lr){5-8} 
\end{tabular}%
%}
%\caption{Fine-tuned DT and RF confusion matrices comparison with all features}
\label{confusion_GB_2}

\\

\begin{tabular}{@{}lrrrrrrrrr@{}}
\cmidrule(r){1-4} \cmidrule(lr){5-8} 
\begin{tabular}[c]{@{}l@{}}DT 20 features\\ max SDS-score\end{tabular} & c & e & o &   \begin{tabular}[c]{@{}l@{}}DT 20 features\\ F-measure\end{tabular} & c & e & o &\\ \cmidrule(r){1-4} \cmidrule(lr){5-8} 
c & 483 & 13 & 3 &   c & 483 & 13 & 3 &    \\
e & 8 & 199 & 3 &   e & 8 & 199 & 3 & \\
o & 24 & 9 & 67 &   o & 24 & 9 & 67 &  \\ \cmidrule(r){1-4} \cmidrule(lr){5-8} 
\end{tabular}%
%}
%\caption{Fine-tuned DT and RF confusion matrices comparison with all features}
\label{confusion_rf_2}

\\

\begin{tabular}{@{}lrrrrrrrrr@{}}
\cmidrule(r){1-4} \cmidrule(lr){5-8} 
\begin{tabular}[c]{@{}l@{}}RF PCA\\ max SDS-score\end{tabular} & c & e & o &   \begin{tabular}[c]{@{}l@{}}RF PCA\\ max F-measure\end{tabular} & c & e & o &\\ \cmidrule(r){1-4} \cmidrule(lr){5-8} 
c & 485 & 9 & 5 &   c & 485 & 9 & 5 &    \\
e & 14 & 194 & 2 &   e & 14 & 194 & 2 & \\
o & 34 & 12 & 54 &   o & 34 & 12 & 54 &  \\ \cmidrule(r){1-4} \cmidrule(lr){5-8} 
\end{tabular}%

%\caption{Fine-tuned DT and RF confusion matrices comparison with all features}
\label{confusion_gb_3}

\\

\begin{tabular}{@{}lrrrrrrrrr@{}}
\cmidrule(r){1-4} \cmidrule(lr){5-8} 
\begin{tabular}[c]{@{}l@{}}DT PCA\\ max SDS-score\end{tabular} & c & e & o &   \begin{tabular}[c]{@{}l@{}}DT PCA\\ max F-measure\end{tabular} & c & e & o &\\ \cmidrule(r){1-4} \cmidrule(lr){5-8} 
c & 483 & 9 & 7 &   c & 483 & 9 & 7 &    \\
e & 25 & 172 & 13 &   e & 25 & 172 & 13 & \\
o & 36 & 11 & 53 &   o & 36 & 11 & 53 &  \\ \cmidrule(r){1-4} \cmidrule(lr){5-8} 
\end{tabular}%

\\

\begin{tabular}{@{}lrrrrrrrrr@{}}
\cmidrule(r){1-4} \cmidrule(lr){5-8} 
\begin{tabular}[c]{@{}l@{}}RF LDA\\ max SDS-score\end{tabular} & c & e & o &   \begin{tabular}[c]{@{}l@{}}RF LDA\\ max F-measure\end{tabular} & c & e & o &\\ \cmidrule(r){1-4} \cmidrule(lr){5-8} 
c & 478 & 9 & 12 &   c & 478 & 9 & 12 &    \\
e & 12 & 191 & 7 &   e & 12 & 191 & 7 & \\
o & 22 & 6 & 72 &   o & 22 & 6 & 72 &  \\ \cmidrule(r){1-4} \cmidrule(lr){5-8} 
\end{tabular}

\\

\begin{tabular}{@{}lrrrrrrrrr@{}}
\cmidrule(r){1-4} \cmidrule(lr){5-8} 
\begin{tabular}[c]{@{}l@{}}DT LDA\\ max SDS-score\end{tabular} & c & e & o &   \begin{tabular}[c]{@{}l@{}}DT LDA\\ max F-measure\end{tabular} & c & e & o &\\ \cmidrule(r){1-4} \cmidrule(lr){5-8} 
c & 482 & 9 & 8 &   c & 482 & 9 & 8 &    \\
e & 17 & 187 & 6 &   e & 17 & 187 & 6 & \\
o & 21 & 7 & 72 &   o & 21 & 7 & 72 &  \\ \cmidrule(r){1-4} \cmidrule(lr){5-8} 
\end{tabular}
%}
%\caption{Fine-tuned DT and RF confusion matrices comparison with PCA}
\label{confusion_rf_3}

\\

\begin{tabular}{@{}lrrrrrrrrr@{}}
\cmidrule(r){1-4} \cmidrule(lr){5-8} 
\begin{tabular}[c]{@{}l@{}}SVM 15 features\\ max SDS-score\end{tabular} & c & e & o &   \begin{tabular}[c]{@{}l@{}}SVM 15 features\\ F-measure\end{tabular} & c & e & o &\\ \cmidrule(r){1-4} \cmidrule(lr){5-8} 
c & 490 & 6 & 3 &   c & 490 & 6 & 3 &    \\
e & 9 & 195 & 6 &   e & 9 & 195 & 6 & \\
o & 25 & 5 & 70 &   o & 25 & 5 & 70 &  \\ \cmidrule(r){1-4} \cmidrule(lr){5-8} 
\end{tabular}%

\\

\begin{tabular}{@{}lrrrrrrrrr@{}}
\cmidrule(r){1-4} \cmidrule(lr){5-8} 
\begin{tabular}[c]{@{}l@{}}DT 15 features\\ max SDS-score\end{tabular} & c & e & o &   \begin{tabular}[c]{@{}l@{}}DT 15 features\\ max F-measure\end{tabular} & c & e & o &\\ \cmidrule(r){1-4} \cmidrule(lr){5-8} 
c & 490 & 5 & 4 &   c & 490 & 5 & 4 &    \\
e & 8 & 197 & 5 &   e & 8 & 197 & 5 & \\
o & 26 & 10 & 64 &   o & 26 & 10 & 64 &  \\ \cmidrule(r){1-4} \cmidrule(lr){5-8} 
\end{tabular}%

\\

\begin{tabular}{@{}lrrrrrrrrr@{}}
\cmidrule(r){1-4} \cmidrule(lr){5-8} 
\begin{tabular}[c]{@{}l@{}}RF 15 features\\ SDS-score\end{tabular} & c & e & o &   \begin{tabular}[c]{@{}l@{}}RF 15 features\\ F-measure\end{tabular} & c & e & o &\\ \cmidrule(r){1-4} \cmidrule(lr){5-8} 
c & 488 & 7 & 4 &   c & 489 & 7 & 3 &    \\
e & 7 & 194 & 9 &   e & 8 & 194 & 8 & \\
o & 22 & 4 & 74 &   o & 22 & 3 & 75 &  \\ \cmidrule(r){1-4} \cmidrule(lr){5-8} 
\end{tabular}%
%\caption{SVM confusion matrices comparison}
\label{confusion_rf_4}

\\

\begin{tabular}{@{}lrrrrrrrrr@{}}
\cmidrule(r){1-4} \cmidrule(lr){5-8} 
\begin{tabular}[c]{@{}l@{}}ET 15 features\\ SDS-score\end{tabular} & c & e & o &   \begin{tabular}[c]{@{}l@{}}ET 15 features\\ F-measure\end{tabular} & c & e & o &\\ \cmidrule(r){1-4} \cmidrule(lr){5-8} 
c & 487 & 8 & 4 &   c & 487 & 8 & 4 &    \\
e & 7 & 198 & 5 &   e & 7 & 198 & 5 & \\
o & 27 & 5 & 68 &   o & 27 & 5 & 68 &  \\ \cmidrule(r){1-4} \cmidrule(lr){5-8} 
\end{tabular}%
%\caption{Gradient booster confusion matrices comparison}
\label{confusion_et_2}
%end{table*}

\\

\begin{tabular}{@{}lrrrrrrrrr@{}}
\cmidrule(r){1-4} \cmidrule(lr){5-8} 
\begin{tabular}[c]{@{}l@{}}GB 15 features\\ SDS-score\end{tabular} & c & e & o &   \begin{tabular}[c]{@{}l@{}}GB 15 features\\ F-measure\end{tabular} & c & e & o &\\ \cmidrule(r){1-4} \cmidrule(lr){5-8} 
c & 488 & 6 & 5 &   c & 488 & 6 & 5 &    \\
e & 8 & 194 & 8 &   e & 8 & 194 & 8 & \\
o & 20 & 5 & 75 &   o & 20 & 5 & 75 &  \\ \cmidrule(r){1-4} \cmidrule(lr){5-8} 
\end{tabular}%
%\caption{Random forest confusion matrices comparison}
\label{confusion_gb_5}

\\

\begin{tabular}{@{}lrrrrrrrrr@{}}
\cmidrule(r){1-4} \cmidrule(lr){5-8} 
\begin{tabular}[c]{@{}l@{}}kNN 15 features\\ SDS-score\end{tabular} & c & e & o &   \begin{tabular}[c]{@{}l@{}}kNN 15 features\\ F-measure\end{tabular} & c & e & o &\\ \cmidrule(r){1-4} \cmidrule(lr){5-8} 
c & 489 & 8 & 2 &   c & 489 & 8 & 2 &    \\
e & 12 & 193 & 5 &   e & 12 & 193 & 5 & \\
o & 30 & 7 & 63 &   o & 30 & 7 & 63 &  \\ \cmidrule(r){1-4} \cmidrule(lr){5-8} 
\end{tabular}%
%\caption{Extra trees confusion matrices comparison}
\label{confusion_knn_2}

\\

\begin{tabular}{@{}lrrrrrrrrr@{}}
\cmidrule(r){1-4} \cmidrule(lr){5-8} 
\begin{tabular}[c]{@{}l@{}}MLP 15 features\\ SDS-score\end{tabular} & c & e & o &   \begin{tabular}[c]{@{}l@{}}MLP 15 features\\ F-measure\end{tabular} & c & e & o &\\ \cmidrule(r){1-4} \cmidrule(lr){5-8} 
c & 485 & 6 & 8 &   c & 485 & 6 & 8 &    \\
e & 9 & 194 & 7 &   e & 9 & 194 & 7 & \\
o & 19 & 5 & 76 &   o & 19 & 5 & 76 &  \\ \cmidrule(r){1-4} \cmidrule(lr){5-8} 
\end{tabular}%
\\
\caption{Confusion matrices for the second experiment (raw data Table~\ref{tab:best_experiment_lda}).}
\label{confusion_table_resum_2}
\end{longtable}

\newpage
\section{Running times of classifiers per 10-fold classification}

\label{appendix-third}

The average running time of extracting one feature of the cell were 2 ms with standard deviation of 0.8 ms.

\begin{table}[ht!]
\centering
\begin{tabular}{|l|l|l|l|}
\hline
\textbf{Classifier}           & \textbf{Features} & \textbf{Average time {[}s{]}} & \textbf{Std time {[}s{]}} \\ \hline
\multirow{2}{*}{\textbf{SVM}} & all               & 3.17       & 1.19           \\
                              & 15                & 0.6        & 0.15    \\ \hline
\multirow{4}{*}{\textbf{DT}}  & all               & 0.35       & 0.56           \\
                              & 15                & 0.07       & 0.04          \\
                              & PCA               & 0.12       & 0.09                  \\
                              & LDA               & 0.08       & 0.01              \\ \hline
\multirow{4}{*}{\textbf{RF}}  & all               & 4.61       & 2.45          \\
                              & 15                & 2.11       & 1.07              \\
                              & PCA               & 1.76       & 0.46          \\
                              & LDA               & 1.46       & 0.69          \\ \hline
\multirow{2}{*}{\textbf{ET}}  & all               & 2.17       & 1.46          \\
                              & 15                & 0.89       & 0.48                       \\ \hline
\multirow{2}{*}{\textbf{GB}}  & all               & 118.05     & 58.29            \\
                              & 15                & 17         & 7.11          \\ \hline
\multirow{2}{*}{\textbf{kNN}} & all               & 6.69       & 4.35                     \\
                              & 15                & 0.63       & 0.55          \\ \hline
\multirow{2}{*}{\textbf{MLP}} & all               & 36.8       & 18.73                \\
                              & 15                & 18.79      & 13.96             \\
                              \hline
\end{tabular}
\caption{Average running times of classifiers (Experiment 1 and Experiment 2), for one iteration grid search cross-validation.}
\label{running_time_table_1}

\end{table}

\begin{table}[ht!]
\centering
\begin{tabular}{|l|l|l|}
\hline
\textbf{Classifier}           &\textbf{Average time {[}ms{]}} & \textbf{Std time {[}ms{]}} \\ \hline
{\textbf{SVM}}                & 30       & 2           \\
                             \hline
{\textbf{DT}}                & 3       & 0.3           \\
                              
                               \hline
{\textbf{RF}}                 & 30       & 2          \\
                             \hline
{\textbf{ET}}                 & 50       & 6          \\                    \hline
{\textbf{GB}}                 & 1068     & 20            \\ \hline
{\textbf{kNN}}                & 2       & 0.2                     \\\hline
{\textbf{MLP}}                & 2391       & 45                \\
                              \hline
\end{tabular}
\caption{Training average running times of classifiers, with 15 features only fitting the classifier with training data without cross validation (Experiment 2).}
\label{running_time_table_2}

\end{table}

%\newpage
%\input{appendixD}

\end{document}